\documentclass[acmtog]{acmart}
\AtBeginDocument{%
  }

\usepackage{bm}
\usepackage{algorithm}
\usepackage{algpseudocode}
\usepackage{siunitx}
\usepackage{amsmath}
\usepackage[table]{xcolor}
\usepackage{hyperref}

\begin{document}

\title{Learning Smooth Time-Varying Linear Policies with an Action Jacobian Penalty}

\acmSubmissionID{385}

\author{Zhaoming Xie}
\email{zxie@rai-inst.com}
\author{Kevin Karol}
\email{kkarol@rai-inst.com}
\author{Jessica Hodgins}
\email{jhk@rai-inst.com}
\affiliation{%
  \institution{Robotics AI Institute}
  \city{Cambridge}
  \state{Massachusetts}
  \country{USA}
}


\begin{abstract}

Reinforcement learning provides a framework for learning control policies that can reproduce diverse motions for simulated characters. However, such policies often exploit unnatural high-frequency signals that are unachievable by humans or physical robots, making them poor representations of real-world behaviors. Existing work addresses this issue by adding a reward term that penalizes a large change in actions over time. This term often requires substantial tuning efforts.
We propose to use the action Jacobian penalty, which penalizes changes in action with respect to the changes in simulated state directly through auto differentiation. This effectively eliminates unrealistic high-frequency control signals without task specific tuning.
While effective, the action Jacobian penalty introduces significant computational overhead when used with traditional fully connected neural network architectures. 
To mitigate this, we introduce a new architecture called a Linear Policy Net (LPN) that significantly reduces the computational burden for calculating the action Jacobian penalty during training. In addition, a LPN requires no parameter tuning, exhibits faster learning convergence compared to baseline methods, and can be more efficiently queried during inference time compared to a fully connected neural network.
We demonstrate that a Linear Policy Net, combined with the action Jacobian penalty, is able to learn policies that generate smooth signals while solving a number of motion imitation tasks with different characteristics, including dynamic motions such as a backflip and various challenging parkour skills.
Finally, we apply this approach to create policies for dynamic motions on a physical quadrupedal robot equipped with an arm.
\end{abstract}

\ccsdesc[500]{Computing methodologies~Animation}
\ccsdesc[500]{Computing methodologies~Control methods}
\ccsdesc[500]{Computing methodologies~Reinforcement learning}

\keywords{Linear Control, Physics-based Character Animation, Legged Robots}


\begin{teaserfigure}
  \includegraphics[width=0.8\linewidth, trim=0cm 6cm 10cm 0cm, clip]{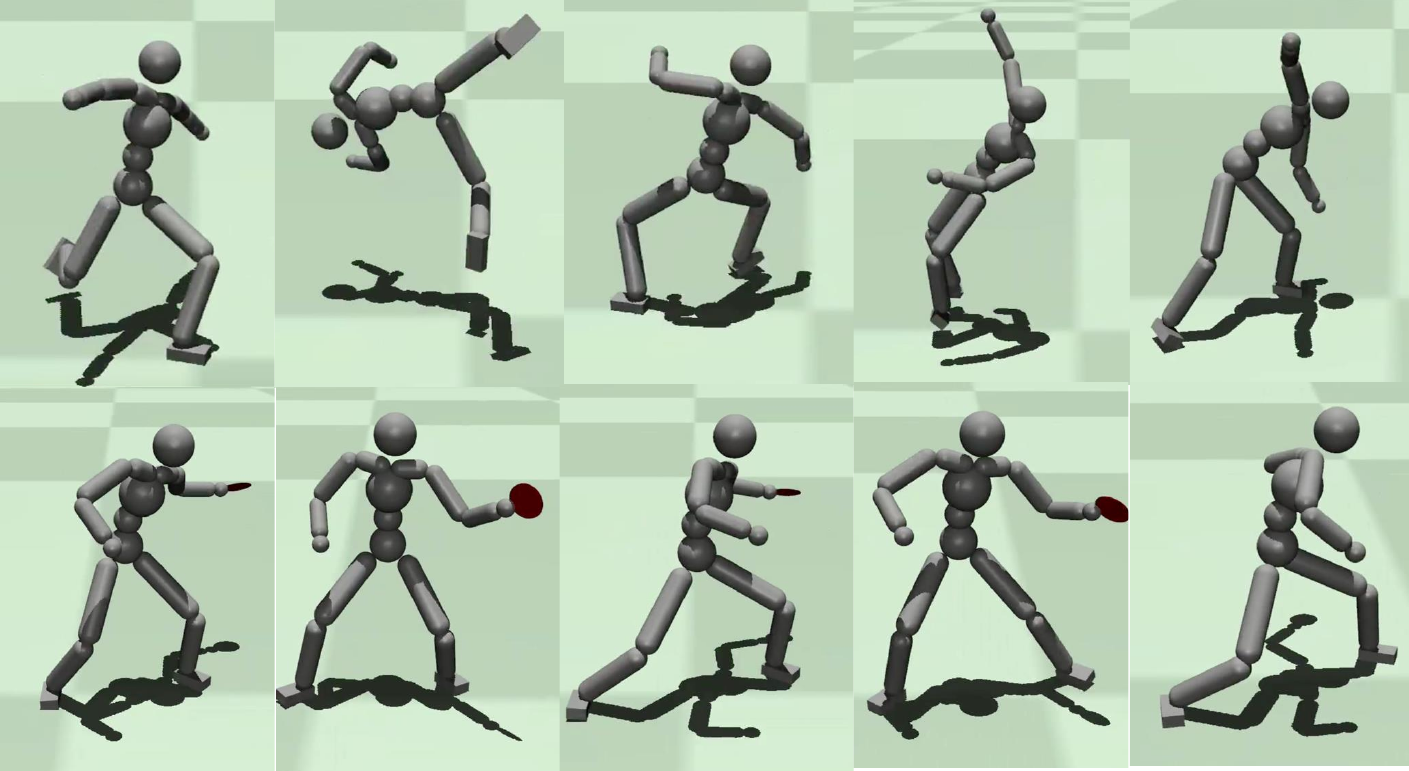}
  \centering
  \caption{Our system is able to learn smooth time-varying linear feedback policies for a simulated character to perform challenging motion skills.
  }
  \label{fig:teaser}
\end{teaserfigure}

\maketitle

\section{Introduction}

Deep reinforcement learning (DRL) has proven effective in physics-based character animation and robotics. However,
policies that naively optimize for task rewards often favor unrealistic, high-frequency control signals. Even when the policies are constrained to imitate high-quality motion data, these unrealistic behaviors still prevail. This problem can be attributed to the learned policies being overly sensitive to slight changes in input signals. Deploying policies that are sensitive in this way on robots is more problematic, because input signals come from noisy sensory measurements and the control bandwidth is limited by what the physical actuators can achieve. Prior work addressed this issue by applying heavy penalties on the rate of change of control signals and randomizing policy inputs to decrease sensitivity to variation. However, these methods are often task-dependent and require significant trial-and-error effort to tune.

More recent work proposes using the Lipschitz-constrained policies to improve the smoothness of a learned control policy. In this method, a gradient penalty is imposed on the likelihood of a control action under the current policy~\cite{lipschitz_policies}. While this method has been effective for locomotion tasks, it remains unclear how well it extends to more challenging scenarios. Furthermore, it relies on a large number of action samples to accurately estimate the true sensitivity of the actions with respect to the inputs and imposes significant computational overhead due to the additional backpropagation required for the gradient penalty.

We propose an action Jacobian penalty, which penalizes the norm of the Jacobian of the control action with respect to the state of the character. This penalty significantly reduces high-frequency signals in learned policies across a variety of motion imitation tasks. However, similar to the Lipschitz-constrained policies, computing the Jacobian of a policy parameterized by a fully connected feed forward (FF) neural network incurs significant computational overhead. To address this, we introduce Linear Policy Net (LPN), a new architecture for parameterizing control policies. Instead of directly outputting a control action, an LPN outputs a feedback control matrix using only task information. This feedback matrix then operates on the state of the character to produce the control action. Computing the Jacobian of the action with respect to the state, and computing the gradient of the Jacobian penalty with respect to the network parameters, is reduced to a simple forward and backward pass of the neural network. This approach effectively reduces the computation time required to impose the Jacobian penalty during training.

In DeepMimic style tasks, learning a LPN is equivalent to learning a time-varying linear feedback policy. This parametrization significantly restricts the class of controller the policy can learn, compared to a standard FF neural network. Our surprising finding is that
this restriction does not negatively impact learning performance or the quality of the final policies. On the contrary, a LPN is comparable to, and in some cases even surpasses, its FF counterparts. We demonstrate the effectiveness of a LPN with a Jacobian penalty on various motion imitation tasks using a simulated human character, including challenging motions such as a backflip, the table tennis footwork drill and challenging parkour skills that require nontrivial interactions with the environment. Furthermore, we show that the learned time-varying linear feedback control policies extracted from a LPN are capable of controlling a physical quadrupedal robot with an arm to perform a combination of dynamic hopping and arm-swing motions.
\section{Related Work}

In this section, we review related work in physics-based character animation with deep reinforcement learning and methods to train smooth control policies. We also review related work in synthesizing linear feedback controllers for simulated characters and robots.

\subsection{Deep Reinforcement Learning}
DRL has become a popular framework for physics-based character animation. Realistic character animations are often synthesized through imitation of kinematic references from motion capture data, e.g.,~\cite{deepmimic, add_siggraph_asia_2025}. An advantage of DRL compared to other approaches is its scalability and versatility. Over the past decade, a wide range of specialized tasks have been tackled with DRL, including scene interaction~\cite{parkour_from_video_siggraph2021}, object manipulations~\cite{skillmimic, intermimic}, multi-character interaction~\cite{multicharacter-siggraph2023}, music instrument play~\cite{piano-siggraph2024} and sports play~\cite{pingpong_siggraph2024, soccer_siggraph2025}. By scaling up the system, hours of diverse human motion capture data can be reproduced in simulation, e.g.,~\cite{scalable_siggaph2020, phc, maskedmimic}. The versatility of this framework also opens up applications such as human motion tracking in virtual reality~\cite{questenvsim} and sim-to-real humanoid control~\cite{beyondmimic, disney_robot_2025, sonic}.

\subsection{Learning Smooth Policies}
Reinforcement learning tends to exploit high frequency signals to achieve high task rewards, which can produce jittery motions that degrade motion quality~\cite{too_stiff} and results in sim-to-real failure for robotics applications. High frequency signals can be reduced by adding a filter to the action~\cite{drecon}, however, this significantly reduces response to perturbation and can impact performance when performing dynamic motions. Directly penalizing the rate of change of actions in the form of a reward signal has been the prevailing approach in sim2real RL for quadrupedal and humanoid robot control~\cite{anymal_rl, spot_rl_2025, beyondmimic}. This approach relies on random exploration of the policy during training to discover smooth behaviors, requires manual tuning to achieve a balance between task completion and behavior regularization. ~\citet{lipschitz_policies, smooth_policies} directly penalized the change of action from the policy, via approximation to the Jacobian of the action with respect to the policy input. While this solution requires less per-task tuning, it imposes significant computational costs resulting in long training times and limiting it to simple control tasks like locomotion.

\subsection{Linear Feedback Control}
Before DRL became popular, control policies were often formulated as linear feedback controllers. Motion can be segmented into stages and a linear feedback controller can be designed for each stage to generate locomotion motion controllers~\cite{simbicon_2007} and various athletic behaviors~\cite{animating_athletes_1995}. These segmented controllers require substantial human effort as each motion needs custom designed segmentations and input features .

Sampling based strategies can be used to search for linear feedback controllers. \citet{reduce_linear_control_SCA2015} learns reduce-order linear feedback controllers via an evolutionary algorithm by directly sampling the feedback matrix. \citet{control_fragment_2016} learns time varying linear feedback controller via iterative sampling and linear regression. These systems are able to learn robust linear feedback policies for dynamic motions such as a backflip and a cartwheel. However, these systems are usually complex, require manual definition of a state representation specific to a given motion, and have not been shown to generalize to motions that require more complex environment interactions such as vaulting and wall climbing.

Through linear matrix parameterization, reinforcement learning or evolutionary algorithms can also be used to train a linear feedback policy. This approach has been shown on various locomotion tasks. These policies either exploit unrealistic behaviors such as high frequency leg motions, e.g.,~\cite{random_search_linear_neurips_2018}, or require carefully designed features, e.g.,~\cite{linear_biped_IROS_2021}. 

Time varying linear feedback control policies can also be synthesized using model-based control approaches such as differential dynamic programming and its variants, e.g.,\cite{ddp_2009, ilqr_2004}. However, linear feedback controllers obtained via model-based approaches are often brittle, and often require online replanning using computationally expensive model predictive control approaches, e.g., ~\cite{mpc_siggraph2019}.

There is also work that applies linearization to the equation of motion. This method allows for an efficient MPC formulation because it only requires solving a small scale Quadratic Programming (QP) problem online. This approach has proven to be effective for quadrupedal and bipedal locomotion, e.g.,~\cite{convex_mpc_2018, linear_humanoid_2025}. However, the resulting controllers are still nonlinear due to the presence of inequality constraints such as the friction cone constraints. 
\section{Problem Setup and System Overview}

Our problem formulation is similar to DeepMimic~\cite{deepmimic}, with a reference motion $\mathcal{M} = \{\hat{\bm{s}}_1, \hat{\bm{s}_2}, \dots, \hat{\bm{s}}_T\}$ provided as input, where $\hat{\bm{s}}_t \in \mathbb R^n$ specifies the full pose of the character state at timestep $t$. A control policy is a map $\pi: \mathbb R^n \times \mathbb R^n \to \mathbb R^m$ where the character state $\bm{s}_t$ and the reference state $\hat{\bm{s}}_t$ are used to generate a control action via $\bm{a}_t = \pi(\bm{s}_t, \hat{\bm{s}}_t)$. The control action $\bm{a}_t$ is a target angle for each actuated joint of the character. A joint level Proportional-Derivative (PD) controller is then used to actuate the character in a physics simulation. The control policy is trained with DRL to drive the character to imitate the reference. In our experiment, we use the same humanoid character as DeepMimic and a simulated quadrapedal robot, modeling the Boston Dynamics Spot quadraped, with arm attached. 

Our reward set up is a simplified version of DeepMimic, where

$$r = 0.3 r_\text{pos} + 0.3 r_\text{orientation} + 0.4 r_\text{joint}$$
with

$$r_\text{pos} = \exp(-50 * \lVert\hat{\bm{x}} - \bm{x}\rVert^2),$$
$$r_\text{orientation} = \exp(-10 * \lVert\hat{\bm{ori}} \ominus \bm{ori}\rVert^2),$$
$$r_\text{joint} = \exp(-2 * \lVert\hat{\bm{j}} - \bm{j}\rVert^2),$$
where $\bm{x}, \bm{ori}$ and $\bm{j}$ are the root position, root orientation and the joint angles of the character respectively. Similar to DeepMimic, we also apply reference state initialization and early termination to improve the learning efficiency.

We use Mujoco~\cite{mujoco} to simlulate the characters. The simulation runs at \SI{120}{\hertz} while the policy is updating the joint target at \SI{30}{\hertz}. A deep neural network is used to parameterize the policy. In this paper, we experiment with both the standard fully connected feedforward (FF) neural net and the Linear Policy Net (LPN) (described in Section~\ref{sec:LPN}). During training, the actions are sampled from a fixed Gaussian distribution where the mean is the output of the policy and the covariance matrix is a diagonal matrix with diagonal element $\delta^2 = 0.01$. 

We use Proximal Policy Optimization (PPO)~\cite{ppo} to optimize the policies. At each iteration of the PPO algorithm, $50$ parallel simulation environments are used to collect $2500$ samples by using the control policy to interact with the simulations. PPO uses these samples to optimize for a loss function
$\mathcal L_\text{PPO}$, via gradient descent on the policy parameters. For all the experiments in this paper, we run PPO for a maximum of $5000$ iterations, which takes around $2.5$ hours on a workstation with $12$ CPU cores for running the simulations and a NVIDIA RTX A6000 for neural network inference and optimization. Most of the time the training already converges after $2000$ iterations, which takes about $1$ hour with $5$ million samples collected. This performance is comparable to a GPU-based simulation framework in terms of both training time and number of simulation samples required, e.g.,~\cite{add_siggraph_asia_2025}.

Additional reward terms are often added to the original DeepMimic loss to reduce motion artifacts such as motion jitteriness or high impact, especially for robotics applications~\cite{beyondmimic, lipschitz_policies}. These additional terms cause additional tuning efforts and often require task specific tuning. In this work, we propose to use a Linear Policy Net with an  action Jacobian penalty as a regularization loss. This approach introduces smooth behaviors to the policies with minimal tuning and compute overhead while also maintaining learning efficiency. We now describe each component in the more detail.
\section{Action Jacobian Penalty}

Reinforcement learning relies on injecting Guassian noise to the policy output for exploration and data collection. This process creates unrealistic high frequency signals that can achieve high rewards. Reinforcement learning then tends to fit the policies to these high frequency signals, resulting in policies that can produce undesired and unnatural jittery control signals. This problem is more pronounced in robotics applications, where real world noise from sensor measurements and imperfect motor commands amplify the high frequency control signals until they are visible. Existing work relies on adding a term in the reward function to penalize the difference between the actions over consecutive time steps to regularize action change over time. But 
the regularization effect tends to be small compared to the other rewards to avoid reducing the effectiveness of the task reward. The size of this effect sometimes causes it to become ineffective in the face of challenging tasks since various reward terms are summed over and the effect of the regularization need to be discovered by the inherent noisy nature of the exploration procedure. 

We propose directly regularizing the policy with an action Jacobian penalty. Instead of using it as a reward signal, we directly add a loss term to the PPO optimization:

$$\mathcal L_\text{total} = \mathcal L_\text{PPO} + w_\text{Jac} \mathcal L_\text{Jac},$$
where $w_\text{Jac}$ is a tunable weighting factor, and $\mathcal L_\text{Jac} = \lVert \mathbf{J} \rVert^2$ is the square of the Frobenius norm of $\mathbf{J}$, with $\mathbf{J}$ the Jacobian of the policy:
\[\renewcommand\arraystretch{1.6}
\mathbf{J} = \begin{bmatrix}
    \frac{\partial a_1}{\partial s_1} & \cdots & \frac{\partial a_1}{\partial s_n} \\
    \vdots & \ddots & \vdots \\
    \frac{\partial a_m}{\partial s_1} & \cdots & \frac{\partial a_m}{\partial s_n}
\end{bmatrix}.
\]
We use $w_\text{Jac} = 10$ across all our experiments.

The action Jacobian captures the sensitivity of the action generated by the policy with respect to the changes in character state. For $\bm{J}$ with a larger norm, a small variation in the character state will cause large changes in action. The resulting motion will exhibit a high frequency joint oscillation. By penalizing the magnitude of the action Jacobian, a policy will generate smoother control signals.

There are prior works that try to approximate $\bm{J}$ to encourage a smooth policy. \citet{smooth_policies} use sampling strategies around the collected data point to approximate the Jacobian, which incurs significant computation costs to get an accurate approximation. \citet{lipschitz_policies} penalizes the gradient of likelihood of a sampled action. This approach is equivalent to penalizing
$\lVert \mathbf{J}^T (\bm{a} - \bm{a}_\text{mean}) \rVert^2$, where $\bm{a}$ is the sampled action during exploration and $\bm{a}_\text{mean}$ is the mean of the Gaussian distribution the policy will sampled from. This penalty only optimizes for a specific direction of the Jacobian and requires more samples to optimize the full Jacobian. We propose directly optimizing for the norm of the full Jacobian, which can be achieved via auto-differentiation and backpropagation.

While penalizing the norm of the action Jacobian is straightforward thanks to the autograd features in a modern deep learning framework such as PyTorch~\cite{pytorch}, it imposes significant computation overhead. Optimizing for the action Jacobian penalty with a fully connected neural network, each iteration of PPO is about $1.5$ times slower in our implementation than using only the PPO loss.
\section{Linear Policy Net}
\label{sec:LPN}
\begin{figure}
    \centering
    \includegraphics[width=0.5\linewidth, trim=8.5cm 6cm 9.5cm 6cm, clip]{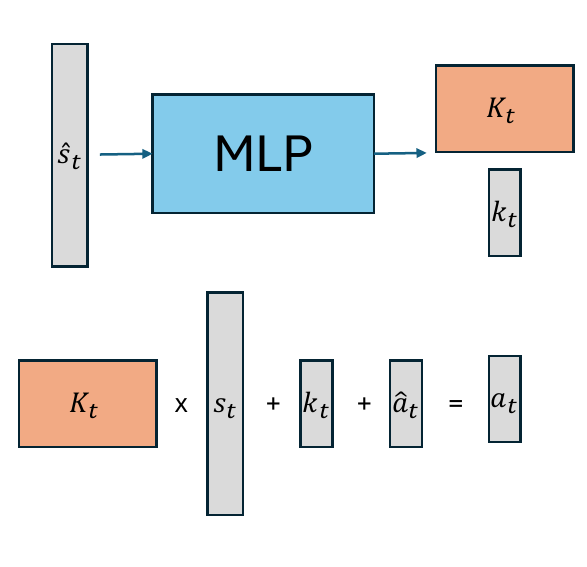}
    \caption{We introduce Linear Policy Net. A fully connected multilayer perceptrons (MLP) is used to generate a feedback matrix $K_t$ and a feedforward action $\bm{k}_t$ from the reference state $\hat{\bm{s}}_t$. The final control action is generated by applying the linear feedback matrix on the simulation state plus the feedforward terms.}
    \label{fig:lpn}
\end{figure}

We introduce a Linear Policy Net (LPN) to parameterize our policies. See Figure~\ref{fig:lpn}. The input to the policy is the tuple $\{\bm{s}_t, \hat{\bm{s}}_t\}$ that represents the state of the simulated character and the desired state from the reference motion at time step $t$. The reference motion $\hat{\bm{s}}_t$ is fed into a two layer MLP, the output of the MLP is a feedback matrix $K_t \in \mathbb R^{m \times n}$ and feedforward action $\bm{k}_t \in \mathbb R^m$, where $n$ is the dimension of the state of the character and $m$ is the dimension of the control action. Control action $\bm{a}_t = \bm{K}_t \bm{s}_t + \bm{k}_t + \hat{\bm{a}}_t$ is then applied to advance the simulation, where $\hat{\bm{a}}_t$ is the reference joint angle corresponding to the actuated joints, extracted from the reference state. Note that $\bm{K}_t$ and $\bm{k}_t$ do not depend on the state of the character, in the case of a fixed sequence of reference motion, this approach is equivalent to learning a time-varying linear feedback control policy, where the feedback matrix and feedforward action only depend on time.

There are several design choices to be made and we will discuss them in this section.

\subsection{Input Features}
Prior work on learning linear feedback controllers often includes manual design features such as the end effector positions in the input~\cite{reduce_linear_control_SCA2015, control_fragment_2016}. DeepMimic similarly includes maximal coordinate of the character that includes positions and orientations of the body parts~\cite{deepmimic}. We choose to use the minimal coordinate system of the character as our policy input, which includes the displacement in the position and orientation of the root to the desired configuration specified in the reference motion, the root linear and angular velocity, and the joint angle and joint velocity of the actuated joints of the character. These features can also be easily obtained from a state estimator on a physical quadrupedal robot, making sim-to-real straightforward.

Another option is to use both the reference state and character state as input to the MLP. While this makes the policy more expressive, the linear feedback matrices would then depend on the character state. We found this dependency unnecessary as the LPN can successfully learn time-varying linear feedback policies for dynamic motion in both simulation and on a physical robot without it.


\subsection{Action Space}

An added benefit of learning a linear feedback matrix is that we can update the feedback matrix and feedforward term $\bm{K}_t, \bm{k}_t$ at a slower rate than $\bm{a}_t$. Such a hierarchy is already present in current DRL frameworks for legged robot control, where the policy is updating the joint target angle at a slow rate of around \SI{50}{\hertz}, while there is a joint level PD controller running at a much higher frequency of \SI{500}{\hertz} or more. Within our framework, we can inference the MLP at even lower rate, down to \SI{15}{\hertz} for some tasks, while maintaining the update frequency of the joint PD target computation and joint level PD control.

Note that because the learnable parameters in the system are in the MLP layer, we can potentially use the feedback matrix $\bm{K}_t, \bm{k}_t$ as our action space, and then do the linear feedback within the simulation loop, hidden from the learning loop. However, this approach increases the action space dimension from $m$ to $(n+1)\times m$, making learning much harder.

\subsection{Action Jaocbian Penalty with Linear Policy Net}

Because the control policy learned by a LPN is in the form of $\bm{a}_t = \bm{K}_t \bm{s}_t + \bm{k}_t + \hat{\bm{a}}_t$, with $\bm{K}_t$ being independent of $\bm{s}_t$, the action Jacobian at time step $t$ is equal to $\bm{K}_t$. Computing the action Jacobian penalty and computing its gradient with respect to the learnable parameters then becomes a simple forward pass and backward pass of the MLP layer in the LPN. This process incurs minimal additional computation cost because the forward and backward pass are already required for evaluating the PPO loss term.
\section{Evaluations}

\begin{table*}[h]
\centering
\caption{Comparison of smoothness of policies trained with different methods, average over three runs with different random seeds. We highlight the best performance. The LPN has top three performance in all the measures except for the motion jerk metric for the backflip task.}
\label{tab:smoothness_metric}
\begin{tabular}{@{}lccccccc@{}}
\toprule
\textbf{Metric by Motion Type} & \textbf{LPN} & \textbf{FF+Jac Pen} & \textbf{Lipschitz} & \textbf{No Reg} & \textbf{reward 0.01} & \textbf{reward 0.1} & \textbf{reward 1} \\ 
\midrule
\rowcolor[HTML]{EFEFEF} \textbf{Walking} & & & & & & & \\
Action Smoothness $\downarrow$  & 0.0016 & \textbf{0.0014} & 0.0040 & 0.0031 & 0.0032 & 0.0025 & 0.0015 \\
High Frequency Ratio $\downarrow$ & \textbf{0.9} & 2.1 & 9.5 & 8.8 & 9.2 & 5.5 & 1.5 \\
Motion Jerk $\downarrow$  & 115.6 & 108.5 & 139.5 & 134.2 & 134.3 & 134.0 & \textbf{106.53} \\
\midrule
\rowcolor[HTML]{EFEFEF} \textbf{Backflip} & & & & & & & \\
Action Smoothness $\downarrow$  & 0.061 & 0.042 & 0.195 & 0.148 & 0.068 & \textbf{0.031} & N/A \\
High Frequency Ratio $\downarrow$ & 8.7 & 4.0 & 33.5 & 26.6 & 11.8 & \textbf{2.8} & N/A \\
Motion Jerk $\downarrow$  & 140.8 & \textbf{109.6} & 170.8 & 168.5 & 132.9 & 111.2 & N/A \\
\midrule
\rowcolor[HTML]{EFEFEF} \textbf{Footwork} & & & & & & & \\
Action Smoothness $\downarrow$ & 0.009 & 0.014 & 0.036 & 0.053 & 0.018 & 0.010 & \textbf{0.005} \\
High Frequency Ratio $\downarrow$ & 1.3 & 5.6 & 21.0 & 27.1 & 13.6 & 2.0 & \textbf{0.5} \\
Motion Jerk $\downarrow$  & 116.6 & 124.6 & 164.6 & 178.1 & 143.5 & 118.0 & \textbf{94.6} \\
\bottomrule
\end{tabular}
\end{table*}

We evaluate our systems on a wide range of physics-based character animation tasks and on a physical quadrupedal robot. In all the examples, the MLP layer in the LPN is a 2-layer fully connected neural network with hidden layer size $256$. For baseline methods, we use the 2-layer fully connected network (FF) that is typically used in these tasks, with hidden layer size 256.

\subsection{DeepMimic Task with Simulated Humanoid}

We apply our method to a number of motion imitation tasks for a simulated humanoid. These tasks can be classified into four categories.

\paragraph{Locomotion Tasks}
We apply our framework to train the humanoid to imitate reference motions for walking and running. We use the reference data from DeepmMimic~\cite{deepmimic}.

\paragraph{Gymnastic Motion}
To demonstrate that our system works on dynamic motions, we train the LPN with the action Jacobian penalty to imitate a range of dynamic gymnastic motions, including a backlip, a sideflip and a cartwheel. While a time-varying linear feedback policy has been shown to work with these motions previously~\cite{control_fragment_2016}, it requires human designed input features that may not generalize to other motions. We demonstrate that simple input features work across all the motions we considered.

\paragraph{Imitating Single Sequence}
We apply our framework to train the humanoid to imitate a 15 second clip of a motion capture of a table tennis footwork drill. The data is obtained from ~\cite{pingpong_siggraph2024}, and involves a dynamic whole body motion where the subject rapidly hops sideways while performing a fast forehand drive. We also demonstrate tracking of a break dance motion, with motion retargetted to the simulated character using motion capture data from~\cite{CMU_Mocap_DB}. This result demonstrates the generalization of our approach beyond cyclic motion.

\paragraph{Environment Interaction}
We demonstrate that our system is also able to learn to interact with the environment with non-trivial contact, such as a parkour motion, with motion capture dataset from~\cite{add_siggraph_asia_2025, parc}. In particular, we train policies to execute a reverse vault sequence, a wall climbing sequence and a double kong sequence, which has been shown to be challenging to learn using DeepMimic~\cite{add_siggraph_asia_2025}. We also learn to imitate a soccer juggling sequence, following the set up from~\cite{soccer_juggling}, to demonstrate the interactions with a dynamic environment.

\subsection{Comparison}
\begin{figure*}
    \centering
    \includegraphics[width=1\linewidth]{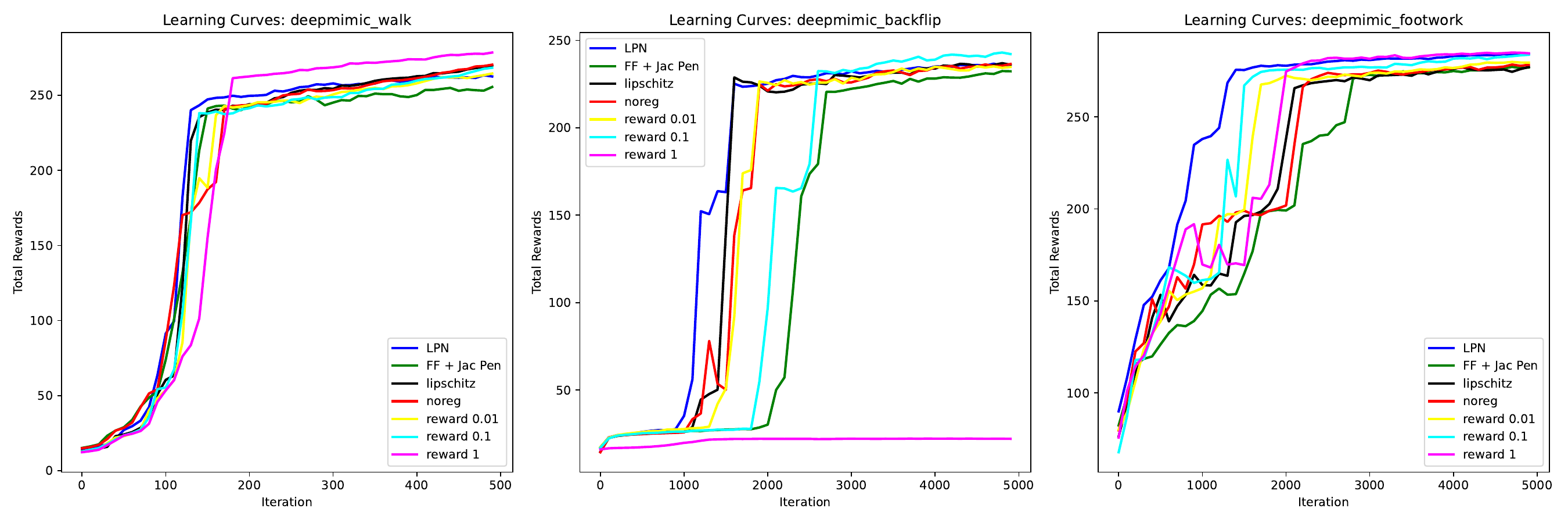}
    \caption{Learning curves of various methods, average over $3$ runs with different random seeds. The total reward is evaluated only on the motion imitation reward.}
    \label{fig:learning_curve}
\end{figure*}

We compare our system with four alternatives, all implemented with FF neural network: 
\begin{itemize}
    \item \textbf{Feedforward Neural Net with an action Jacobian Penalty}, where we apply the action Jacobian penalty during training. This is labeled as \textbf{FF + Jac Pen}.
    \item \textbf{No regularization}, where we directly optimize for $\mathcal L_\text{PPO}$ using the imitation reward. This is labeled as \textbf{No Reg}.
    \item \textbf{Action Change Penalty}, where a reward $r_\text{action} = -w_\text{action}\lVert \bm{a}_t - \bm{a}_{t-1} \rVert^2$ is used to penalize the changes in action between two timesteps. The weight $w_\text{action}$ is a tunable parameter. We experiment with three sets of weights: $w_\text{action} = 0.01, 0.1, 1$. We label them as \textbf{reward 0.01}, \textbf{reward 0.1} and \textbf{reward 1}.
    \item  \textbf{Lipschitz Constraint Policy~\cite{lipschitz_policies}}, where instead of minimizing the norm of the action Jacobian, a penalty is applied to $\mathcal L_\text{Lipschtiz} = \frac{\lVert \bm{a} - \bm{a}_\text{mean} \rVert^2}{ds}$, where $\bm{a}_\text{mean}$ is the mean of the Gaussian distribution the policy will sample from while $\bm{a}$ is the sampled action. This method is equivalent to minimizing  $\lVert\bm{J}^T(\bm{a} - \bm{a}_\text{mean})\rVert^2$. It does not require computation of the whole Jacobian matrix but will only optimize in the direction of the sampled $\bm{a}$. We use the same weighting factor as the action Jacobian penalty and optimize for $\mathcal L_\text{total} = \mathcal L_\text{PPO} + 10 \mathcal L_\text{Liptschitz}$. This is labeled as \textbf{Lipschitz}.
\end{itemize}
We compare the learning performance on walking, a backflip and the table tennis footwork drill, taking the average over three training runs with different random seeds. The learning curve is shown in Fig.~\ref{fig:learning_curve}. The reward is computed only with the imitation reward. Applying a large action penalty (\textbf{Reward 1}) slows down task learning but results in better motion imitation performance, except for the backflip, where a large action change penalty causes learning to fail. FF with an action Jacobian Penalty takes almost twice as many learning iterations to converge compared to other methods, and each iteration is about $1.5$ time slower due to the Jacobian penalty computation. The Lipschitz constraint policy shows fast convergence in walking and backflip tasks, but converges slowly for the footwork task. Furthermore, it fails to produce a smooth policy compared to the alternatives, as we will show later. The LPN with the Jacobian penalty has the fastest learning convergence across all tasks with minimal computation overhead per iteration.

To quantify the smoothness of a policy, We collect data by rolling out the policy in simulation. For walking and the backflip, we run the policy for five motion cycles and for the table tennis footwork drill, we run the policy over the complete sequence. We also quantify the smoothness of each policy using the following metrics:
\begin{itemize}
    \item \textbf{Action Smoothness}: This metric evaluates the average action change of a policy: $\frac{\sum^{T+1}_{t=1}\lVert \bm{a}_t - \bm{a}_{t-1} \rVert^2}{T}$.

    \item \textbf{High Frequency Ratio}: We compute the Fourier transform of the action output over time. Humans typically have a control bandwidth of about \SI{10}{Hz}~\cite{biomechanics_bandwith}, we consider signal content higher than \SI{10}{Hz} to be unnatural. We use the proportion of the energy that is higher than \SI{10}{Hz} with respect to the total energy to characterize the smoothness of the signal. 

    \item \textbf{Motion Jerk} We compute the jerk metric of the motion, following ~\cite{jerk_metric}. From the joint acceleration signal sampled at \SI{120}{Hz}, we use finite differencing to compute the joint jerk. We then evaluate the jerk metric by dividing the mean jerk magnitude by the peak speed, average over the $28$ joints on the character. A lower value in the jerk metric corresponds to a smoother motion. 
\end{itemize}
The result is recorded in Table~\ref{tab:smoothness_metric}. Heavily penalizing the action change in the reward (\textbf{reward 1}) can produce smooth policies. However, it fails to learn dynamic tasks like a backflip. \textbf{reward 0.1} is also able learn smooth behaviors, with a reduced leaning convergence rate, especially for the backflip task. Task specific reward tuning is necessary to obtain the best performance with the action change penalty. The Lipshcitz constraint policies consistently fail to achieve smooth policies compared to the other methods. FF policies with an action Jacobian penalty are able to learn smooth policies that are competitive with the methods that use a reward that heavily penalizes action change, but they are slow to train. A LPN is able to learn smooth policies while maintaining fast learning convergence rate. Note that in the backflip task, the smoothness metric for a LPN is worse than the feedforward neural network policies with a Jacobian penalty or appropriate reward set up. We conjecture that the backflip is a challenging motion for a time-varying linear feedback control policies, requiring the LPN to produce higher frequency action.

\begin{figure*}
    \centering
    \includegraphics[width=1\linewidth]{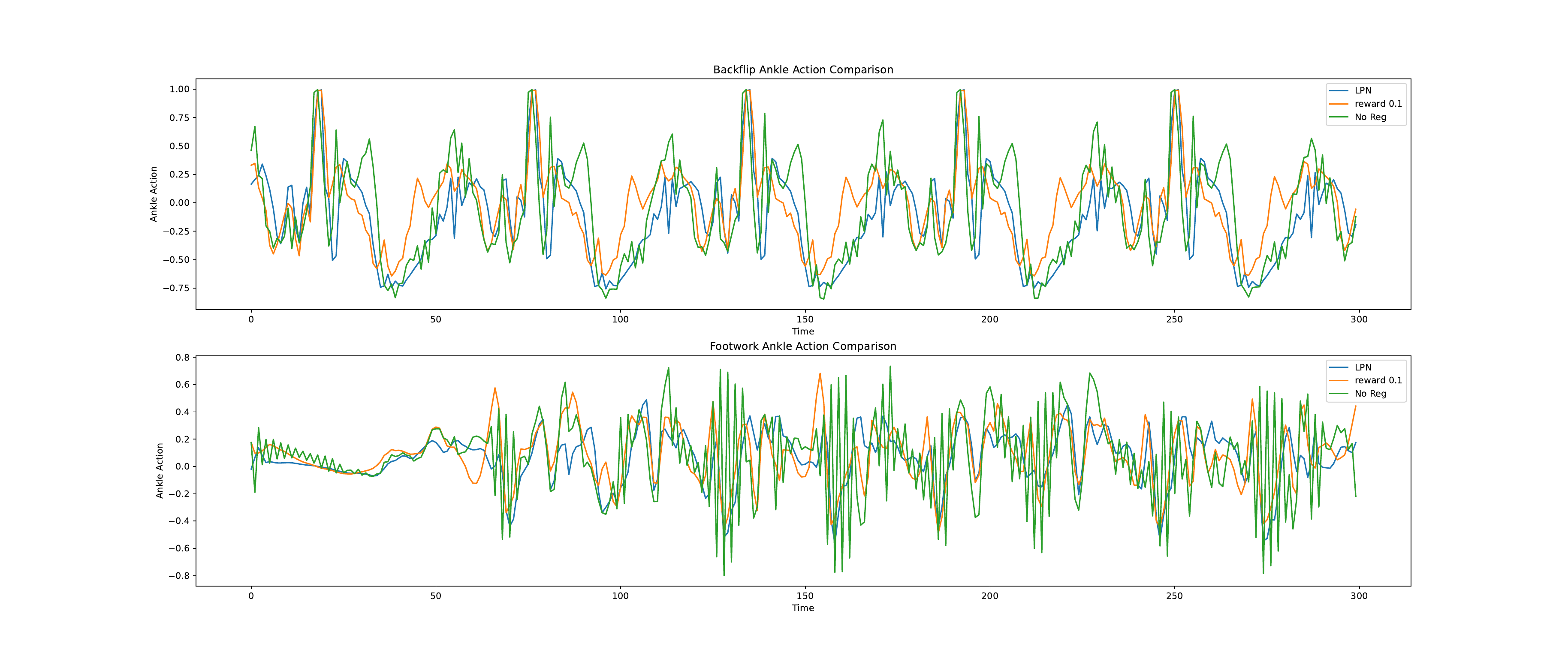}
    \caption{Control action of the pitch joint on the left ankle for the backflip and table tennis footwork drill.}
    \label{fig:ankle_action}
\end{figure*}

We also plot the action of the pitch joint on the left ankle for the backflip and table tennis footwork in Fig.~\ref{fig:ankle_action} to provide a visual demonstration of the importance of regularizing the control action, either via an action change penalty as a reward signal or using our action Jacobian penalty. Without action regularization, policies tend to generate actions that change rapidly, resulting in jittery motions, potentially causing the feet to chatter on the ground.

\subsection{Linear Policy Net Evaluations}

\paragraph{Reduced-Order Linear Feedback Policy} Inspired by~\cite{reduce_linear_control_SCA2015}, we experiment with how to obtain a reduced-order linear policy by using low rank linear feedback matrices. We perform singular value decomposition on the learned feedback matrices to compute their low rank approximation. Specifically, for a feedback matrix $\bm{K}_t$, its singular value decomposition is in the form of $\bm{K}_t = \bm{U}\bm{\Sigma}\bm{V}^T$, with $\bm{U}$ and $\bm{V}$ being orthogonal matrices that form bases for the action space and state space of the character respectively, and $\bm{\Sigma}$ is a diagonal matrix, whose diagonal elements quantify the importance of the corresponding dimension in those bases. A best rank $k$ approximation of the feedback matrix can be obtained via $\bm{K}_{k, t} = \bm{U}_k \bm{\Sigma}_k \bm{V}_k^T = \sum_{i=1}^{k} \sigma_i \bm{u}_i \bm{v}_i^T$, where $\bm{u}_i$ and $\bm{v}_i$ are the $i$th column of $\bm{U}$ and $\bm{V}$ respectively.

Our action space is $28$ dimensional corresponding to the $28$ joints on the character. We find that we can lower the rank of the feedback matrices learned by a LPN while still maintaining performance. For example, a sequence of rank $14$ feedback matrices are able to retain the performance of a walking policy. We can further reduce the rank of these matrices down to two. While they can still command the character to walk, the motion quality degrades.

We can also find low rank approximation of the policies for other motions. For example, the rank of a backflip policy can be reduced to $20$, the rank of a cartwheel policy can be reduced to $22$, and the rank of a table tennis footwork drill policy can be reduced to $18$. 

\paragraph{Terrain Adaptation}  We adapt our backflip policy and cartwheel policy to handle uneven terrain by finetuning them on a sinusoidal terrain. Even though these policies do not perceive the terrain, the underlying linear feedback structure is able to handle the perturbation. This experiment demonstrates the robustness of the time varying linear feedback polices learned by a LPN.

\paragraph{Linear Feedback Update Rate} LPN is trained to update the linear feedback matrix and the feedforward action at a rate of \SI{30}{Hz}. We experiment with how much we can slow down the update rate while still maintaining good motion tracking performance. We find that we can update the feedback matrix at \SI{10}{Hz} with a walking policy, while policies for other motions fail when we try to lower the update rate below \SI{30}{Hz}. In ~\cite{control_fragment_2016}, the linear feedback control policies can be queried at a slow rate of \SI{10}{Hz}, even for highly dynamic motions such as a backflip. It will be interesting to figure out how to train a LPN to operate at this slower rate.

\paragraph{Policy Distillation and Transitions between Skills} We can train policies for different motions and distill them into a single policy, following the procedure in~\cite{pdp}. The resulting policies can track different reference motions in sequence. In particular, we train a LPN policy to track jumping, a sideflip and a backflip via distillation. The resulting policies can then execute jumping, a sideflip and a backflip in sequence. We also train a FF policy via the same distillation procedure, it fails to perform the agile transition between a sideflip and a backflip.

\subsection{Sim-to-real on a Quadrupedal Robot}
We demonstrate that the time-varying linear feedback policies trained with our framework can be readily applied to a physical robot. We use a quadrupedal robot Spot with an arm attached to the body as our platform. To reduce the sim-to-real gap, we implement the actuator model of the leg motors that correspond to the physical actuator model specified by the manufacturer, following~\cite{spot_rl_2025}. The results are shown in the supplementary video.

To train a locomotion policy, a reference motion is generated for a pacing gait on Spot using a handcrafted sinusoid for the joints on the legs, with a gait cycle of 0.6 second, similar to~\cite{dynamics_randomization_icra2021}. During training, the joints on the arm are set to a random target sampled around the current configuration every $0.5$ second. The result is a time-varying linear feedback policy that can maintain stable pacing motion while executing fast arm movement. Instead of querying the LPN during run time, we precompute a sequence of linear matrices offline and apply them in sequence. We update the linear feedback matrices at \SI{15}{Hz}, and compute the joint target angles for the PD controller using linear feedback at \SI{30}{Hz}. This significantly reduces the computation load required to maintain stable motions with a FF policy, which requires inference of a neural network at \SI{50}{Hz}~\cite{spot_rl_2025}.

To demonstrate a combination of an agile arm movement and an agile lower body movement, we generate a hopping motion via trajectory optimization of a single rigid body model~\cite{convex_mpc_2018}, and an agile table tennis stroke motion for the arm using a kinematic MPC planner~\cite{spot_pingpong_2025}. After synchronizing these motions, we create a reference motion for Spot that is similar to the table tennis footwork practice drill of a human player. Again, an LPN is able to learn a time-varying linear feedback policy to execute the motion.

\section{Conclusions and Discussion}

In this paper, we present a framework to efficiently train smooth time-varying linear control policies for motion imitation tasks for a simulated character and a physical quadrupedal robot.

A surprising finding is that this time-varying linear policy exists for a wide range of motions without needing special feature engineering. Another way to synthesize a time-varying linear feedback policy is via model-based control such as differential dynamic programming (DDP), although matrices obtained via DDP are often less robust. It would be interesting to explore how to combine both approaches, so that we can gain the benefit of sample efficiency from model-based approaches while retaining the robustness afforded by the DRL framework. For example, we could warm start the feedback matrices with solutions from the DDP method, e.g.,~\cite{guided_policy_search}. Given that our our learned policies are in the simple form of linear matrices, the policies can potentially be more easily explained than a typical black box feedforward neural network policy. For example, it is possible that there exists a DDP formulation that with the appropriate costs and transition dynamics can reproduce the same feedback matrices learned by our system. Inverse optimal control techniques can potentially be used to search for this formulation, e.g., using~\cite{differentiable_mpc}.

While the action Jacobian penalty produces smooth policies for many motion imitation tasks, it only considers the derivatives with respect to state but not time. For dynamic motions such as a backflip, where the state of the character has to change rapidly, penalizing the action Jacobian alone is not guaranteed to reduce the changes of action in time. This observation may also explain why using the action Jacobian penalty is less effective for the backflip in our experiment.

As a future work, it will be interesting to explore how we can learn piece-wise linear policies in the state space. For example, we can segment the state space into regions where there exists a linear control policy for each region. Neural network policies that use ReLU as an activation function can already be used to generate such regions based on the activation patterns~\cite{relu_tedrake, linear_region}. However, without regularization, such regions are usually too fine-grained. Learning linear feedback control policies that have a large region of attractions allowing us to divide the state space into manageable pieces might further improve policy robustness and explainability.

We focus on the imitation of relatively short motion segments compared to current DRL systems that can scale to imitating long motion sequences. It is conceivable that by scaling the system to imitate more and longer motions, one may find a one-to-one correspondence between feasible motion capture data and the set of feedback matrices. Such correspondence can enable us to learn a policy generator, where a generative model such as a diffusion model~\cite{policy-space_diffusion} can be used to generate the linear feedback policies. 

While we demonstrate skill composition via policy distillation, we are not yet able to transition between arbitrary skills. Scaling the system to a large motion dataset and building a control graph~\cite{control_fragment_2016} can potentially automate more diverse transitions.  

Our formulation limits our use case to DeepMimic style motion imitation tasks. Expanding the formulation to other tasks, e.g., using adversarial motion imitation~\cite{add_siggraph_asia_2025}, or tasks where motion capture data is not available will be an interesting direction.


\bibliographystyle{ACM-Reference-Format}
\bibliography{sample-base}

\appendix

\end{document}